\documentclass[runningheads]{llncs}
\pdfoutput=1
\usepackage{graphicx}
\usepackage{cite}
\usepackage{amsmath,amssymb,amsfonts}
\usepackage{algorithmic}
\usepackage{textcomp}
\usepackage{xcolor}
\usepackage{comment}
\usepackage{epstopdf}
\usepackage{scalerel,stackengine}

\begin{document}

\title{A Subword Level Language Model for Bangla Language}

\titlerunning{Subword Language Model}

\author{Aisha Khatun\inst{1}\orcidID{0000-0001-6559-4493} \and
Anisur Rahman\inst{1}\orcidID{0000-0002-4616-4559} \and
Hemayet Ahmed Chowdhury\inst{1}\orcidID{0000-0002-5582-669X}\and
Md. Saiful Islam\inst{1}\orcidID{0000-0001-9236-380X}\and
Ayesha Tasnim\inst{1,2}\orcidID{0000-0002-7143-3255}
}

\authorrunning{A. Khatun et al.}

\institute{Department of Computer Science and Engineering, Shahjalal University of Science and Technology\\
Kumargaon, Sylhet 3114, Bangladesh\\
\email{\{aysha.kamal7,emailforanis,hemayetchoudhury,saif.acm\}@gmail.com} 
\and
\email{tasnim-cse@sust.edu}
}

\maketitle

\begin{abstract}
Language models are at the core of natural language processing. The ability to represent natural language gives rise to its applications in numerous NLP tasks including text classification, summarization, and translation. Research in this area is very limited in Bangla due to the scarcity of resources, except for some count-based models and very recent neural language models being proposed, which are all based on words and limited in practical tasks due to their high perplexity. This paper attempts to  approach this issue of perplexity and proposes a subword level neural language model with the AWD-LSTM architecture and various other techniques suitable for training in Bangla language. The model is trained on a corpus of Bangla newspaper articles of an appreciable size consisting of more than 28.5 million word tokens. The performance comparison with various other models depicts the significant reduction in perplexity the proposed model provides, reaching as low as 39.84, in just 20 epochs.

\end{abstract}

\section{Introduction}
There are generally two types of language models, continuous-space\cite{continuous} and count-based\cite{countbased} language models. Count-based models calculate the n-gram probabilities to make n-th order Markov assumptions by counting the frequency of n-gram occurrences, followed by various smoothing techniques. Most of the work in Bangla\cite{sabir} is based on this approach. But n-gram models are extremely sparse. Besides, this method of using n-grams to represent models is linguistically uninformed, because it has to rely on exact matches of patterns of tokens. Furthermore, the language model is limited to a specific window, whereas humans can exploit a much larger context much easily.

Even though language models are at the heart of NLP, the newest techniques of language modeling have not yet been adapted for Bangla. Specifically in the reign of the continuous-space language model, very little has been done for Bangla language. Feed-forward neural language models \cite{continuous} and recurrent neural language models (RNNs) have been implemented to solve the sparsity problems of n-gram models. Here words are represented as vectors of continuous values called word embeddings and these vector representations are fed into the network for training. These vectors maintain the property that semantically close words are are closer in the vector space they are in. Neural Language models have also been used in various other levels of embedding including sentence, corpus, and subword.

In this paper, we use a variant of the recurrent neural language model,  Average-Stochastic-Gradient-Descent Weight-Dropped LSTM\cite{merity}, on subword tokenized Bangla text, which was tokenized using Googles Sentencepiece tokenizer. We used the unigram subword tokenization\cite{kudo2018subword} to tokenize the corpus. This process of tokenization helps provide structural information about the language. Especially for a language with inflection\cite{howardsubword}, such as Bangla. We also present a structure that includes various techniques to optimize the training of the language model, producing significantly low perplexities on data sets.

\section{Related Works} \label{related}

\subsection{On Bangla Language}
Bangla is the seventh most spoken language by the total number of speakers in the world, yet lacks standard resources. Works on modern language modeling are next to existent. Most published works are those of count-based models\cite{sabir}, which have been far outperformed by modern neural networks. Some considerable work was done in \cite{bangla_1}. Besides this, a comparative analysis was performed for word embeddings \cite{ourpaper}. But the need for more improved language models remains. Notably, no language model was trained keeping in mind the level of inflection that occurs in Bangla language, thus tokenizing in subword levels may offer greatly towards downstream tasks such as text classification, summarization, and translation.

\subsection{On Language Models}
Language modeling can be defined as the conditional probability of observing a sequence of words and predicting the next word in the sequence. Traditionally this was accomplished using word-count based methods, which are vastly replaced by neural language models at present. A description of count-based and neural language models is presented below.
\subsubsection{Count-based language models}
N-gram language models are the basic type of count-based methods based on Markov assumption. This model states that the probability of occurrence of a word depends on the last n-1 words that occur before it. A sentence or a paragraph is predicted word by word in this manner. The Markov chain rule states that :
\begin{equation}
p( w_{n}  \mid  w_{1} , w_{2} , \ldots , w_{n-1} ) \approx p( w_{n}  \mid  w_{n-m} , \ldots , w_{n-2}, w_{n-1} )
\end{equation}

Here, `m` is the number of context words.
The number of previous words/history is the order of the model. The basic idea is to predict probability of $w_{n}$ from its preceding context. If the context contains only $w_{n-1}$, it is bi-gram, and the result is obtained by dividing the frequency of $w_{n-1},w_{n}$ by frequency of $w_{n-1}$. Uni-gram on the other hand considers only the frequency of $w_{n}$. Following a similar pattern, a tri-gram model would be as follows: 

\begin{equation}
p(w_{3} \mid w_{1},w_{2}) =  \frac{count(w_{1},w_{2},w_{2})}{ \sum{}_w count(w_{1},w_{2},w)}
\end{equation}

Although a simple concept, n-gram models have several drawbacks. First, out-of-sample combinations are given zero probability. This is called the sparsity problem and tends to occur very frequently. An attempt to combat this is to use back-off \cite{back-off} and smoothing techniques \cite{smooth1}. Second is the curse of dimensionality. Due to the enormous number of possible word combinations, it becomes increasingly unfeasible to train on a larger number of context words.\par

Besides this, the semantic and syntactic similarity of sequences is not recognized by such models, making it unable to `learn` language. True Markov assumption is also not modeled due to the fixed length of context words taken into consideration.

\subsubsection{Continuous-space language models}
Continuous-space language model, also called neural language model(NLM), is the solution to sparsity and high dimensionality problems mentioned previously. These models can represent distributed values and thus learn syntactic and semantic features of the language, which may be otherwise impossible to extract.\par
Recent works exploit NLM to achieve state-of-the-art performance using recurrent neural networks or its variations such as LSTM, along with other optimization and regularization techniques. Neural language models can be of various architectural backgrounds as we see below:

\begin{itemize}
    \item Feed-Forward Language Models:
\end{itemize} \par
Feed-Forward models learn the probability of the next word for the previous n-1 words with the help of a few feed-forward layers. An architecture of this form contains a mapping from each word of the vocabulary V, to a continuous vector space $\in R^{m}$, where m is the feature dimension. Therefore the mapping C is a matrix of size  $\mid V\mid \times m $, each row representing a feature vector of a word. Then a composite function is formed with mappings and a function for calculating conditional probability of $w_t$ with respect to $(w_{t-n+1},\ldots , w_{t-1})$, which
learns the features and parameters.
It can generalize better than the n-gram models and solves the sparsity problems, however, long training and testing times pose a considerable threat to the scalability of such methods. 

Two models were proposed to combat this issue. One\cite{morin2005hierarchical} works on clustering similar words for calculation reduction by building a binary hierarchical tree on the vocabulary words using expert knowledge, while another\cite{mnih2009scalable} uses data-driven methods to do the same. 11.1\% perplexity reduction was achieved from the best HLBL model\cite{mnih2009scalable} comparing with the Kneser-Ney smoothed 5-gram LM.

\begin{itemize}
    \item Recurrent Neural Language Models:
\end{itemize}\par

In recurrent neural network based model(RNNLM)\cite{recurrentLM}, there is no limitation to the context size, so the information can keep circulating within the network for as long as needed. These models provide better generalization by the recurrently connected neurons, also called short term memory.

An improvement over RNNLM proposed in \cite{RLMextensions} implements factorization on the output layer using classes, thus reducing the computational complexity of RNNLM. This version is faster to train and test, and also performs better than the original RNNLM.

\subsection{On Neural Network Architectures}
Recently numerous models have been proposed with more advanced structures and strategies to improve on language modeling. Although most provide good performance, AWD-LSTM\cite{merity}, with its specialized optimization and regularization techniques has been outperforming others in most cases. AWD-LSTM stands for ASGD(Average Stochastic Gradient Descent) Weight-Dropped LSTM. It uses DropConnect\cite{dropconnect} as a means of regularization, a variant of Average-SGD (NT-ASGD) and some other techniques in a very effective way.

LSTM follows a method to utilize short and long term memory of context texts, which can be represented using these generic equations:

\begin{equation}
i_{t} =  \sigma ( W^{i} x_{t} +  U^{i} h_{t-1})
\end{equation}
\begin{equation}
f_{t} =  \sigma ( W^{f} x_{t} +  U^{f} h_{t-1})
\end{equation}
\begin{equation}
o_{t} =  \sigma ( W^{o} x_{t} +  U^{o} h_{t-1})
\end{equation}
\begin{equation}
\widetilde{c}_{t}  =  tanh ( W^{c}x_{t} + U^{c}h_{t-1})
\end{equation}
\begin{equation}
c_{t} =  i_{t}  \odot  \widetilde{c}_{t} +  f_{t} \odot  +  \widetilde{c}_{t-1}
\end{equation}
\begin{equation}
h_{t} =  o_{t}  \odot  tanh(c_{t})
\end{equation}

\begin{figure}[h!]
\includegraphics[scale=0.4]{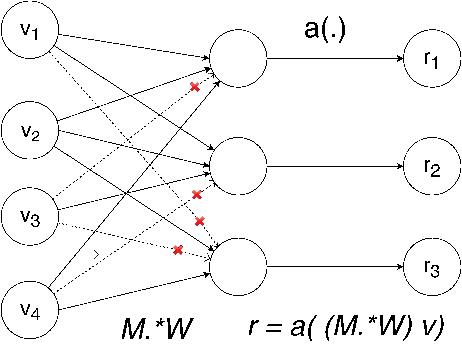}
\centering
\caption{DropConnect Network\cite{dropconnect}}
\label{Fig: DropConnect}
\end{figure}

 Here, $W^{i}, W^{f}, W^{o}, W^{c}, U^{i}, U^{f}, U^{o}, U^{c}$ represents weight matrices, $x_{t}$ is the t-th time step vector input, $h_{t}$ is the present hidden state, $c_{t}$ is the memory cell state, and $\odot$ is element-wise multiplication.
 \newline
 The DropConnect comes in the scene to eliminate the conventional over-fitting caused by RNN/LSTM layers. It is applied on the hidden-to-hidden activations, $(U^{i}, U^{f}, U^{o}, U^{c})$, randomly. This prevents over-fitting, still maintaining RNNs long-term dependency retention. Figure \ref{Fig: DropConnect} provides an illustration of the DropConnect network.

Unlike SGD, ASGD takes into account previous iterations in updates and returns an average value. AWD-LSTM uses a variant of ASGD called NT-ASGD(non-monotonically triggered ASGD), where ASGD is activated if the validation metric does not improve for a fixed number of cycles.
\newline
Several other methods for effective learning includes variation in backpropagation-through-time length, variational dropout\cite{variable_dropout}, and embedding Dropout\cite{variable_dropout}.
These techniques improve the overall structure of the network, making it easier to train a language model effectively, with significantly small model size, and therefore is the first choice of architecture.

\subsection{On Training Neural Networks}

Various methods and techniques can be employed while training a model to make it faster, more effective and perform better. When these techniques meet the right architecture, state-of-the-art performances can be achieved. Some of such methods used in this paper are described below.

Learning rates are an important and yet tedious hyper-parameter when training a model. Finding the right learning rate is extremely crucial, but has been solved over the years only through time expensive trial and error methods. An impressive approach by Leslie Smith\cite{clr} addresses this problem by performing a single trial run over the data. Training starts with a small learning rate and exponential increase occurs per batch. The loss is recorded for every point. The optimum learning rate is the highest one with a descending error.

Another training problem is local minima. Neural networks may often move towards the local minima through gradient descent, instead of global minima, and get stuck. According to \cite{SGDR}, if the learning rate is abruptly increased, the gradient descent will be able to jump out of local minima and start looking for the global minima again. Experiments from \cite{SGDR} prove the effectiveness of this approach. 

\section{Corpus}
\begin{figure}[h!]
\includegraphics[scale=0.6]{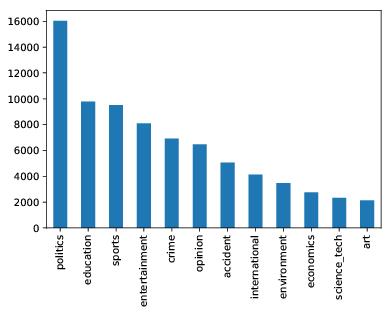}
\centering
\caption{Distribution of classes in the dataset}
\label{datasetfig}
\end{figure}

A large corpus on Bangla newspaper articles was created using a custom web crawler containing 12 different topics. This corpus is relatively large compared to the ones used for language modeling in the past for Bangla\cite{sabir}. The total number of word tokens in this dataset is 28533646 (28.5+ million), with unique word count of 836509. The number of unique words is around 3\% of the entire vocabulary of the dataset. A summarized statistics of the dataset is given in Table \ref{corpustable}. The Dataset is imbalanced, which is apparent from Figure \ref{datasetfig}, for the train set. 20\% of the dataset was separated as a held-out dataset for testing purposes.

\begin{table}[h!]
\centering
\caption{Dataset Statistics}
\begin{tabular}{|c|c|c|c|c|} 
\hline
category & samples & total word & unique word\\
\hline
opinion	& 8098 & 4185472 & 243968\\
international & 5155 & 1089780 & 86852\\
economics & 3449 & 909648 & 58932\\
art & 2665 & 1312571 & 154869\\
science & 2906 & 697899 & 76755\\
politics & 20050 & 6167418 & 196541\\
crime & 8655 & 2016342 & 128308\\
education & 12212 & 3963695 & 225348\\
sports & 11903 & 3087029 & 174677\\
accident & 6328 & 1086791 & 77171\\
environment & 4313 & 1347509 & 103783\\
entertainment & 10121 & 2669492 & 204902\\
\hline
average & 7988 & 2377803 & 144342\\
total & 95855 & 28533646 & 836509\\
\hline
\end{tabular}
\label{corpustable}
\end{table}

\section{Methodology}

\subsection{Proposed Architecture}

This paper employs an architecture called AWD-LSTM\cite{merity} for language modeling. It is a variant of Recurrent Neural Network with LSTM gates\cite{gers} along with special regularization techniques and optimization which allows for context generalization and language understanding. The LSTM layers allow for the memory of the context of a sentence but are prone to over-fitting. AWD-LSTM employs dropconnect, among other techniques making it the choice of interest for language modeling for a language such as Bangla. 

\begin{figure}[h!]
\includegraphics[scale=0.3]{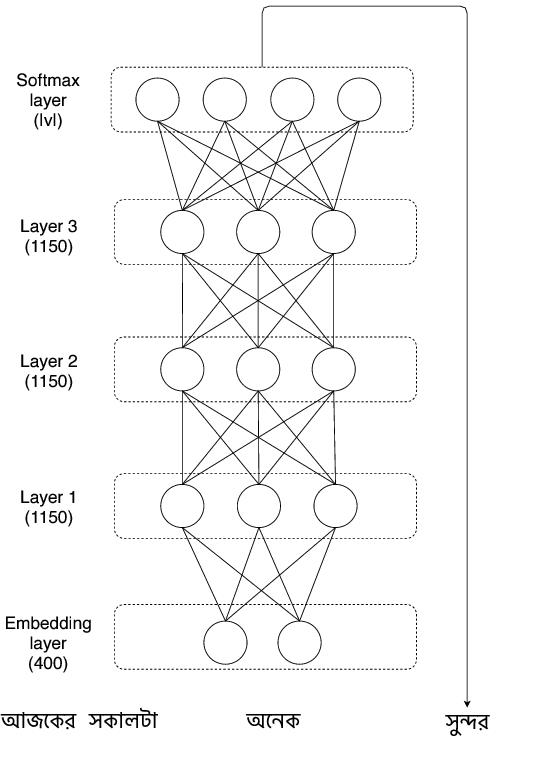}
\centering
\caption{Language Model Architecture}
\label{Fig:awd_bangla}
\end{figure}

This architecture consists of an embedding layer of size 400, followed by 3 regular LSTM layers with no attention. Additionally, it has some short-cut connections and numerous drop-out hyper-parameters. Each layer has 1150 activation neurons. At last a softmax layer provides probabilities for the next word. Adam optimizer is used along with flattened categorical cross-entropy loss function. The architecture is illustrated in figure \ref{Fig:awd_bangla}.

\subsection{Subword Tokenization}
In traditional training, words are the tokens of the model with a fixed vocabulary, where each token is represented by an integer. However, while mapping words to integers we lose information about word structure, therefore the language model requires more time and data to learn about languages with inflections such as Bangla. To help the model learn at a deeper level, one possibility is the use of characters as tokens. But this approach produces much larger models and increases the computational costs many-fold, yet not retaining as much information as required in the LSTM layers\cite{bojanowski2015alternative}. So we opt for a method in the middle, and perform subword tokenization of the text using SentencePiece tokenizer. This tokenizer performs unsupervised tokenization directly from raw sentences irrespective of the language. The unigram segmentation algorithm\cite{kudo2018subword} was employed to create the subword vocabulary.

\subsection{Training the language model}

The SentencePiece tokenizer was set to a vocabulary size of 30,000\cite{howardsubword}, which included \textless s\textgreater,\textless/s\textgreater,\textless unk\textgreater tokens as the start token, end token, and the unknown token. All tokens appearing less than 3 times were discarded and replaced with the \textless unk\textgreater token. The language model was trained with a batch size of 32, back-propagation-through-time window was set to 70. A dropout multiplier of 0.5 was used for all the LSTM layers for regularization. Weight decay of 0.1 was used.

\begin{figure}[h!]
\fbox{\includegraphics[scale=0.6]{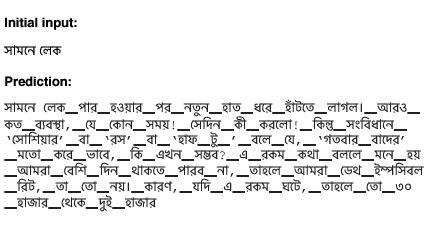}}
\centering
\caption{Example Predictions from the Language Model}
\label{examples}
\end{figure}

The layers were unfrozen, and weights were randomly initialized. Using the cyclical learning rate finder technique\cite{clr}, a learning rate of 1e-3 was selected and the model was trained for 20 epochs. Stochastic gradient descent with restarts (SGDR) mentioned in Section \ref{related} was applied to each epoch.
Given a phrase, the language model at this point was capable of completing entire paragraphs in Bangla language as illustrated in figure \ref{examples}.

\section{Experiments}
We attempt to run some experiments for the sake of comparison, recreating from previously proposed models and evaluate their perplexities. Perplexity is a common measure of language model performance. It calculates how well a model predicts the next possible token, given a context of any length. In this case, perplexity is a way to determine the uncertainty in text generation through probability assignment. Low perplexity is a sign that the model is performing well, whereas high perplexity means the language model was not able to capture the language well. In simple words, perplexity can be calculated as the exponentiation of the loss. The models used for analysis are:

\subsection{Bi-gram language model}
 
 We compare our model with the bi-gram language model following \cite{JelMer}, which has been frequently used for language modeling in Bangla with variations. 
 Let $q_{t} =  freq( w_{t-1},w_{t-2} )$ be the representation of the frequency of occurrence of a specific window of words $(w_{t-1},w_{t-2})$. Therefore the conditional probability takes the following form:
 \begin{equation}
P ( w_t  | w_{t-1}  ) =   \alpha _{0}( q_t ) p_0 +
\alpha _1( q_t ) p_1 ( w_t ) +  \alpha _2( q_t ) p_2 ( w_t | w_{t-1}  )
\end{equation}
 
 With conditional weights $\alpha _{i}( q_{t} ) \geq  0 ,  \sum{}_i \alpha _{i}( q_{t} ) = 1  $.
 Prediction is done using various window size, including the following ways:
 $p_{0} = 1 /|V| , p_{1}$ is a uni-gram and $p_{2}(i|j)$ is the bi-gram. Rest of the experimental measures are kept similar to \cite{sabir}.

\subsection{LSTM and CNN language models}
In order to compare the proposed architecture with its modifications against simpler plain models, we conduct experiments with a few other continuous space neural models. We build a simple LSTM model with two layers that are trained for 10 epochs. Each layer has 200 hidden units and all hidden states are initialized by 0. We train with a batch size of 32, and an initial learning rate of 1 is used.
Besides this, we also try a character level CNN model, similar to the classification model of \cite{character}. The character set is of size 85 containing all Bangla letters, digits among others. The model consists of 2 convolutional layers followed by a fully connected layer. The length of the input sequence is set as 1000 characters, other parameters are kept consistent with \cite{character}.

\subsection{AWD-LSTM word-level language model}
For a much closer comparison, a strand of research by Chowdhury et al, word-level language model using AWD-LSTM architecture was also compared. This model was fed with text by only separating the words as tokens. Vocabulary size of 60,000 was used and all words with frequency less than 3 were eliminated and replaced with \textless unk\textgreater. The model was trained for 4 epochs with a learning rate of 1e-2, and then 3 more epochs with a learning rate of 1e-3 after which the model started to overfit, so training was stopped. This model yielded a perplexity of 51.2 at the end of training on the held-out test dataset.

\section{Results and Discussion}

The held-out test set is evaluated and the perplexities are recorded in the Table \ref{resulttable}.

\begin{table}[h!]
\centering
\caption{Perplexity Comparison of the Language Models}
\begin{tabular}{|c | c|} 
 \hline
 Model & Perplexity on Test Set\\ [0.5ex] 
 \hline\hline
 AWD-LSTM subword(proposed)  & \textbf{39.84}\\ 
 \hline
 AWD-LSTM word & 51.2\\ 
 \hline
 Simple LSTM & 227\\
 \hline
 Character CNN & 125\\ 
 \hline
 Bi-gram & 860.1\\[1ex] 
 \hline
 
\end{tabular}
\break

\label{resulttable}
\end{table}

From the table, it is clear that the subword level model significantly outperformed all other models. The simple LSTM model gained a perplexity of 227 while using AWD-LSTM architecture lowers it up to 39.84. Character level CNN improves the perplexity from simple LSTM model but it is still outperformed by the proposed model significantly. 
The bi-gram model did not work well at all, which shows the need for using a neural language model. Furthermore, the inflection of Bangla language was leveraged by SentencePiece tokenization providing a subword text structure, which when input to the model improves the model performance from 51.2 to 39.84 perplexity.

\section{Conclusion}
In this paper, we discuss the ways to effectively train a weight-dropped LSTM language model in Bangla using subword tokenization and other strategies. We showed that the proposed subword level model outperforms the word-level models with a perplexity of 39.84, showing the effectiveness of such tokenization. This is a significantly low perplexity achieved in Bangla language so far. The reasons for such results are we believe the architectural makeup of AWD-LSTM using DropConnect mask on the hidden-to-hidden weight matrices as a technique of regularization, which is a common problem in language modeling. Besides this, the structure of Bangla language makes it difficult to perform modeling tasks, which we attempted to solve using subword tokenization. By using subwords instead of words, not only does it bring together the same words with various extensions, but it also significantly reduces the vocabulary size required to train the model to a similar level of accuracy. Language models are being used widely for transfer learning in NLP and a well-trained model that understands the language in question will be effectively helpful in downstream tasks such as text summarization, classification, and translation.

\section{Acknowledgement}
This paper is a grateful recipient of the facilities and research environment of the Department of Computer Science and Engineering, Shahjalal University of Science and Technology (SUST) and SUST NLP Research Group. All authors of this paper deserve equal credit providing major contributions to the research.

\vspace{12pt}

\bibliographystyle{splncs04}
\bibliographystyle{unsrt}
\bibliography{biblio}
\end{document}